\documentclass[letterpaper, 10 pt, conference]{ieeeconf}  

\IEEEoverridecommandlockouts                              

\RequirePackage{scrlfile} 
\PreventPackageFromLoading{subfig}
\usepackage{cite}
\usepackage{amsmath,amssymb,amsfonts}
\usepackage{bm}
\usepackage{bbm}
\usepackage{algorithmic}
\usepackage{graphicx}
\usepackage{textcomp}
\usepackage{xcolor}
\usepackage{subcaption}
\usepackage{verbatim}
\usepackage{tikz}
\usepackage{breqn}
\usepackage{booktabs}
\usepackage{multirow}
\usepackage{svg}
\usepackage{hyperref}
\usepackage{lipsum} 

\usepackage{array}
\usepackage{balance}
\usepackage[linesnumbered,ruled,vlined,algo2e]{algorithm2e}
\usepackage{xcolor}

\title{\LARGE \bf
Online Stochastic Variational Gaussian Process Mapping for Large-Scale SLAM in Real Time
}

\author{Ignacio Torroba$^{1}$  Marco Cella$^{1}$ Aldo Terán$^{2}$ Niklas Rolleberg$^{2}$ and John Folkesson$^{1}$
\thanks{This work was supported by Stiftelsen for Strategisk Forskning (SSF) through the Swedish Maritime Robotics Centre (SMaRC) (IRC15-0046)}
\thanks{The authors are with $^{1}$ the Robotics Perception and Learning Division and $^{2}$ the Naval Architecture Division at KTH Royal Institute of Technology, SE-100 44 Stockholm, Sweden{\tt\footnotesize \{torroba, mcella, aldot, nrol, johnf\}@kth.se}}
}

\begin{document}

\maketitle
\thispagestyle{empty}
\pagestyle{empty}

\begin{abstract}
Rao-Blackwellized particle filter (RBPF) SLAM solutions with Gaussian Process (GP) maps can both maintain multiple hypotheses of a vehicle pose estimate and perform implicit data association for loop closure detection in continuous terrain representations. Both qualities are of particular interest for SLAM with autonomous underwater vehicles (AUVs) in the open sea, where distinguishable features are scarce. However, the applicability of GP regression to parallel, real-time mapping in an RBPF framework remains limited by the size of the survey and the computational cost of the GP training. To overcome these constraints, in this letter we propose the adaption of Stochastic Variational GP (SVGP) regression to online mapping in combination with a novel, efficient particle trajectory storing in the RBPF. We show how the resulting RBPF-SVGP framework can achieve real-time performance in an embedded platform on two AUV surveys containing millions of points. We further test the framework on a live mission on an AUV and we make the implementation publicly available.

\end{abstract}
\section{INTRODUCTION}
Autonomous underwater vehicles (AUVs) are becoming standard tools for underwater exploration and seabed mapping in both scientific and industrial applications \cite{graham2022rapid, stenius2022system}. Their capacity to dive untethered allows them to reach areas inaccessible to surface vessels and to collect data more closely to the seafloor, regardless of the water depth. However, their navigation autonomy remains bounded by the accuracy of their dead reckoning (DR) estimate of their global position, severely limited in the absence of a priori maps of the area and GPS signal. Global localization systems equivalent to the later exists for the underwater domain, such as LBL or USBL. However they involve expensive external infrastructure and their reliability decreases with the distance to the AUV, making them unsuitable for deep sea surveys. 

\begin{figure}[h]
    \centering
    \includegraphics[width=\linewidth]{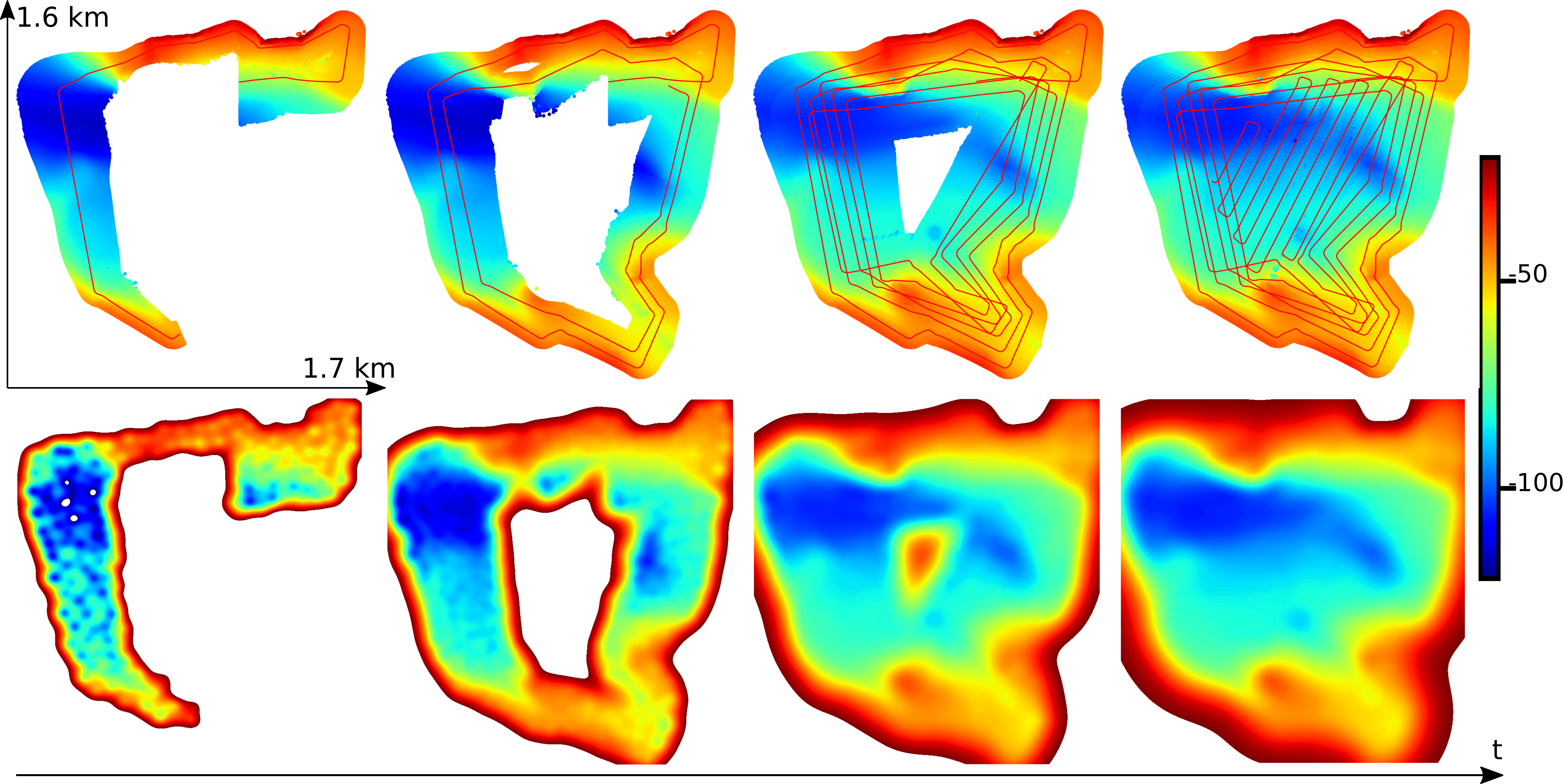}
    \caption{Time-lapse of 1 out of 100 RBPF-SVGP particles learning a bathymetric survey of 8M points in real time.}
    \label{fig:time_lapse}
\end{figure}

Simultaneous localization and mapping (SLAM) techniques can ameliorate these limitations by allowing the vehicle to autonomously adjust its DR estimate based on corrections against a map of the area being built at survey time. 
For bathymetric mapping, the de facto sensor in industry is the multibeam echosounder (MBES), which directly produces 3D representations of the seabed. Thus, the current state of the art SLAM solutions for deep sea navigation target AUVs equipped with an MBES. Two examples of these are \cite{torroba2019towards}, with a maximum a posteriori (MAP) approach to solving SLAM and \cite{barkby2011bathymetric}, based on a Rao-Blackwellized Particle Filter (RBPF) solution. The later methodology presents two advantages when compared to submap-based graph SLAM, as in the former. These two strong points of the RBPF are: i) its capacity to keep several hypotheses of the vehicle pose and ii) its ability to handle the data association problem implicitly. This second point is particularly important in bathymetric SLAM with MBES, where autonomous data association remains a pervasive challenge due to the scarcity of distinguishable features \cite{tan2022data}.
These reasons motivated the recent work in \cite{krasnosky2022massively}, which presented an RBPF solution for bathymetric SLAM based on Gaussian Process (GP) models of the terrain. The use of GP regression (GPR) to represent bathymetry yields continuous and light maps with a readily available measure of map uncertainty. However, the challenge in the use of parallel GPR in online and large scale surveys in embedded hardware arises from their training complexity being cubic on the number of training inputs \cite{rasmussen2003gaussian}. To overcome this limitation, \cite{krasnosky2022massively} presented a GPU-targeted implementation of the GPR for a faster and more efficient learning. However, despite improvements in speed, their framework still required the bathymetry to be discretized a priori into tiles, each with their own GP, and it is unsuitable for embedded platforms.

To enable handling of surveys of millions of points without discretizing them while accomplishing real-time, in water, performance integrated on an AUV system on its payload computer, we propose in this work an RBPF solution based on Stochastic Variational Gaussian Process (SVGP) \cite{hensman2013gaussian} maps. Combining an adaption of SVGP to online mapping with a novel implementation of the handling of particles lineages, we show how our RBPF framework achieves real-time performance with datasets of millions of MBES data points in a NVIDIA Jetson Orin, with up to $100$ particles. We assess the capabilities of the presented SLAM solution in three surveys collected with two AUVs, one of them live, and we make the framework publicly available to encourage further research.


\section{RELATED WORK}
The Rao-Blackwellized factorization of SLAM, first proposed in \cite{montemerlo2002fastslam}, divides its posterior into two separate problems, the vehicle localization and the map construction. Thus, the localization estimate can be computed via a particle filter, in which each particle represents a weighted hypothesis of the AUV pose \cite{torroba2022fully}. Each particle also maintains its own map, from which expected measurements can be computed. By comparing them to the real ones, those particles whose maps resemble the real one more closely are resampled.

Regarding the filter structure, an efficient implementation of the particles' maps is proposed in \cite{barkby2011featureless} based on the Distributed Particle Mapping (DPM) from \cite{eliazar2005hierarchical}. Each particle's trajectory is stored as a succession of the trajectory segments of all its ancestors, for which a so-called ancestry tree is built. Then, a common log of MBES measurements is shared across all particles in synchrony with their trajectories, such that a particle's map can be reconstructed backtracing through the ancestry tree and transforming the measurements into the particle's trajectory. A similar distributed approach has been adopted in this work.

Regarding the map representation, \cite{fairfield2007real} employed 3D evidence grid maps, a type of Octree structure that permitted to accomplish real-time RBPF SLAM in underwater tunnels. Targeting open sea bathymetry instead, in \cite{barkby2011featureless} a 2D grid-based map is used to represent seabed depths. 
GP was originally used to regress seabed models within an RBPF framework in \cite{barkby2011bathymetric} with the approach later extended in \cite{krasnosky2022massively}. Both works modelled bathymetric maps online through GPR and showed the advantages of using GP prediction to perform loop closures in unmapped areas. However, they both suffered from their inability to model the full survey with a single GP. In \cite{barkby2011bathymetric}, the problem was sidestepped by retraining the GP on the latest patch of collected bathymetry at determined time intervals. This method, however, can suffer from "memory loss" in the kernel hyperparameters and therefore might not be able to handle big loop closures. Instead, in \cite{krasnosky2022massively} the survey was divided into patches or tiles of predetermined size. Although effective, this divide-and-conquer approach involved discretizing the space similarly to grid-based mapping, hindering the continuous representation that GP is capable of. Furthermore, it implied equally fixed training times per GP, regardless of the bathymetry in the tile, data redundancies in the overlapping areas across tiles and the need to keep track of which GP to query when computing expected measurements.

\cite{barkby2011bathymetric, krasnosky2022massively} remained therefore limited by the cubic complexity of the GP training on the number of training inputs $N$ \cite{rasmussen2003gaussian}. Other application-based solutions to adapt GP to large-scale mapping tasks \cite{vasudevan2009gaussian} as well as sparse approximations to the GP kernel \cite{quinonero2005unifying} have been presented in the literature. However, in this work we propose adopting the SVGP formulation from \cite{hensman2013gaussian} to the online terrain reconstruction problem in order to handle large datasets efficiently. 

\section{SVGP BATHYMETRIC MAPPING}
This section introduces large-scale bathymetric modelling with SVGP in an offline context.

\subsection{Bathymetric surveying with a MBES}
Autonomous bathymetric surveying tasks itself with creating a map of an unknown area of seabed $E$ gathering MBES data with an AUV.
The DR of the AUV can be modelled as a Gaussian distribution $r_{t} \sim \mathcal{N}(g(r_{t-1}, c_{t}), W)$, where $r_t$ represents the AUV pose estimate at time $t$. $c_t$ and $g()$ are the vehicle's control input and its motion model respectively and $W$ is the covariance of the zero-mean, additive Gaussian noise that parameterizes the uncertainty in the DR. 
Similarly, the MBES pings can be modelled as $P_{t} \sim \mathcal{N}(h(r_t, E_t), Q)$, where $P_t = \{p_{i}\}_{i=1}^{n}$ is a ping containing $n$ 3D beams $p_i = [x, y, z]$, $h()$ is the MBES measurement model and $Q$ parameterizes the measurements uncertainty. $E_t = \{e_{i}\}_{i=1}^{n}$ describes the patches of seabed, in the map frame, where the beams hit the ground at time $t$. 

After a bathymetric survey, a dataset of the form $D = \{p^{xy}_i, p^{z}_i\}_{i=1}^{N}$ can be constructed, containing the $N$ beams collected in time $T$. In the reminder of this paper, we will alternatively use the notation $D = \{X, Y\}$ and the former when necessary to adapt to the standard notation in the GP literature.

\subsection{Stochastic Variational Gaussian Process maps}
If the seabed is assumed to be an unknown continuous function of the form $f: X \to Y$ from which the MBES beams are noisy samples, SVGP can be used to regress the original $f$ from the dataset $D$. Applying Bayes rule, such regression problem takes the following form:
\begin{equation}
\label{eq:bayesian_svgp}
    p(f(X) | Y) = \frac{p(Y | f(X)) p(f(X))}{p(Y)}
\end{equation}
where we seek to determine the posterior distribution that best fits the observations collected.
The prior in Eq. \ref{eq:bayesian_svgp} can be modelled through a GP as $p(f(X)) \sim  \mathcal{N}(0, \mathcal{K}_{NN})$, parameterized be a zero mean and a kernel matrix $\mathcal{K}_{NN}$.
For a selected GP kernel, the optimal set of its hyperparameters $\theta_{kernel}$ can be learned maximizing the likelihood of collecting the N observations in $D$ for the given model. 
However, such maximization involves the inversion of the $\mathcal{K}_{NN}$ matrix, which has historically prevented GP from being used in large mapping applications. In order to alleviate this constraint, several sparse formulations have been proposed \cite{quinonero2005unifying} based on approximating the kernel via a set of $S << N$ latent variables, normally referred to as inducing variables $u$. These inducing variables correspond to a set of input locations $u = f(Z)$. In our problem $Z$ belongs in the same domain as $X$. Furthermore, in order to handle intractable posteriors in Eq. \ref{eq:bayesian_svgp}, a Variational inference (VI) approach  was proposed in \cite{titsias2009variational} to approximate them instead through a user-defined variational term $q()$. Minimizing the KL divergence between the target posterior and the selected $q()$, $KL [q(f(X), u)|| p(f(X),u|Y)]$, is equivalent to maximizing its VI evidence lower bound (ELBO), which yields the optimal set of hyperparameters $\theta_{kernel}$.
However, optimizing such ELBO remains subject to manipulating the $N$ training points, limiting its applicability. To overcome this, the SVGP formulation \cite{hensman2013gaussian} decouples the process complexity from the number of training points by parameterizing the variational term as $q(f(X),u) = p(f(X) | u)q(u)$, with $q(u) \sim \mathcal{N}(\mu, \Sigma)$. This results in the ELBO in Eq.\ref{eq:elbo_svgp}, which can be factorized into $N$ independent terms (we represent $f(X)$ as $f$ in the following for the sake of space): 
\begin{equation}
\label{eq:elbo_svgp}
    \mathcal{L} = \sum^N_{i}\mathbb{E}_{q(u) p(f | u)} \left[\ln p(y_i \mid f_i)\right] - \mathrm{KL}\left[q(u) \mid\mid p(u)\right]
\end{equation}
Thanks to this factorization, stochastic gradient descent (SGD) techniques can be applied to optimize the ELBO in Eq. \ref{eq:elbo_svgp} through minibatch estimates of size $M << N$.
Such minibatch estimator has the form:
\begin{equation}
\label{eq:elbo_mini_svgp}
    \mathcal{\hat{L}} = \frac{N}{M} \sum^M_{i}\mathbb{E}_{q(u) p(f | u)} \left[\ln p(y_i \mid f_i)\right] - \mathrm{KL}\left[q(u) \mid\mid p(u)\right]
\end{equation}
The ELBO in Eq. \ref{eq:elbo_mini_svgp} can be derived with respect to the locations of the $S$ inducing points $Z_s$ and therefore these can be added as hyperparameters in the training process. Therefore the final set becomes $\theta_{SVGP} = \{\theta_{kernel}, Z_s\}$.
Thus, the SVGP formulation has permitted learning offline map models from datasets containing millions of points, as in \cite{torroba2022fully, miller2022mapping}.

\section{THE RBPF-SVGP ALGORITHM}
An RBPF algorithm approximates the SLAM posterior by maintaining an estimate of the vehicle pose through a finite set $R_t$ of $J$ weighted particles, each containing its own map estimate $SVGP^j_t$:
\begin{equation}
    \label{eq:particles}
    R_t = \{\langle r^j_t, SVGP^j_t \rangle\}_{j=1}^{J}
\end{equation}
In order to be able model the particles maps with SVGP, in this section we present our approach to adapt SVGP to online bathymetry learning and its efficient integration in an RBPF framework to achieve real-time management of up to $100$ particles in an embedded platform. 

\subsection{Online mapping with SVGP}
In the previous section, the focus has been on offline SVGP mapping. In order to adapt it to online surveying, two minor modifications need to be adopted involving the construction of the dataset $D$ and the initialization of the inducing locations $Z_s$.

Regarding the dataset, so far we have assumed $D$ to have a fixed size $N$ and contain all the data collected from a survey of duration $T$. However, in order to implement online SVGP training, the dataset must be increased with every new sample collected during the survey. Thus, we reformulate our dataset as $D_t$, to denote its increasing size over time. We will show in our experiments how the SGD minibatch estimators can handle a growing dataset provided a sufficiently high rate of SVGP training iterations per new MBES added to $D_t$.

Concerning $Z_s$, their initial positions must now be selected carefully to account for the evolution of $D_t$. Although the learning of $Z_s$ will eventually result in their optimal positions being found across the dataset, this process is generally slower than the rate at which new data is collected with a moving vehicle. Therefore, the results will be suboptimal if all the $Z_s$ locations are naively initialized over the data in $D_t$ when the survey starts. We overcome this by uniformly distributing $Z_s$ a priori over the area to survey, whose boundaries are known at the time of planning the mission.

\subsection{SVGP minibatch construction}
\begin{figure}[t]
    \centering
    \vspace*{3mm}
    \includegraphics[width=\linewidth]{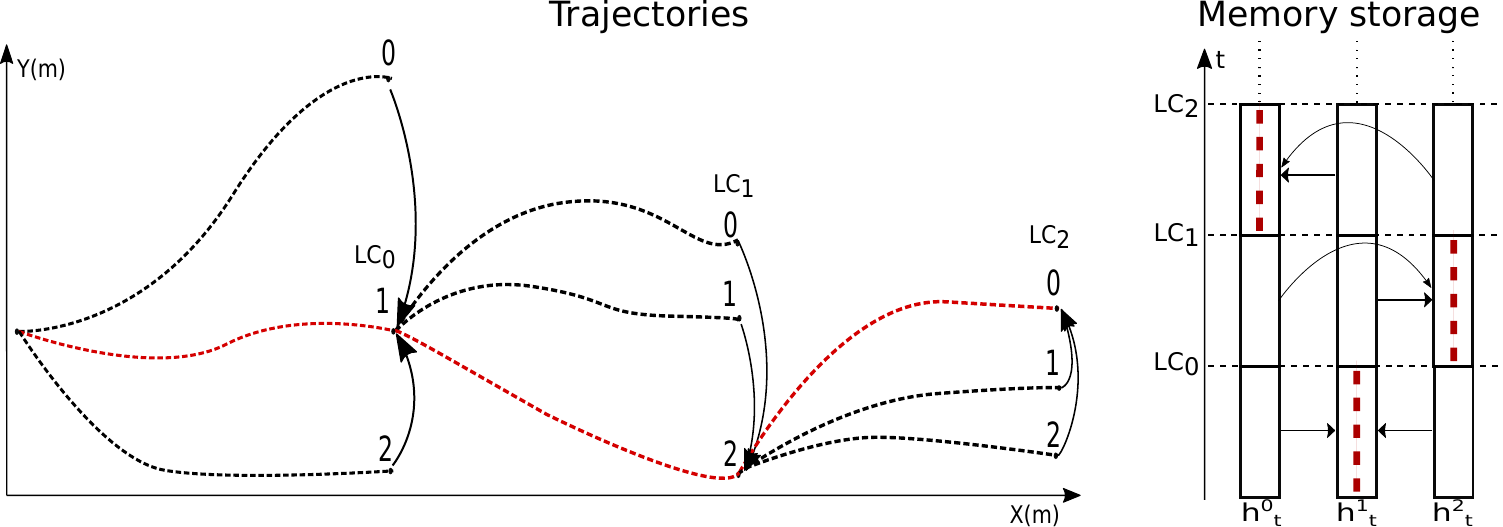}
    \caption{Schematic of the proposed storing of particles trajectories as vectors of pointers to segments of trajectories in between resamples.}
    \vspace*{-3mm}
    \label{fig:particle_mem}
\end{figure}

In an RBPF framework with SVGP maps, constructing minibatches for Eq. \ref{eq:elbo_mini_svgp} entails randomly sampling from both the MBES log $D_t$ and a particle's history of trajectories $h_t^j = \{ r_t^j\}_{t=0}^{t}$, with both of them growing over time.
\begin{algorithm2e}[h]
\caption{$SVGP_j$ training step} \label{algo:minibatch}
\small
    \SetKwFunction{FMain}{svgp_iteration}
    \SetKwProg{Fn}{}{:}{}
    \Fn{\FMain{$D_t, h^j_t, \theta^j_{SVGP,s-1}$}}{
        $D_M = \emptyset$ \\
        \For{\texttt{$m=0$ to $M$}}{
            $ \langle p_{b}, t_b \rangle \sim \mathcal{U}(D_t)$ \\
            $r^j_{tb} \leftarrow h^j_t(t_b)$ \\
            $p^j_{b} = T(r^j_{tb}) \: p_b$\\
            $add$ $p^j_{b}$ $to$ $D_M$ \\
        }
        $\theta^j_{SVGP,s} \gets OptimizerStep(D_M, \theta^j_{SVGP,s-1})$ \\
     \KwRet{$\theta^j_{SVGP,s}$}
    }
\end{algorithm2e}
Algorithm \ref{algo:minibatch} describes the necessary steps involved in an SVGP training iteration for particle $j$. Lines $3-7$ construct a minibatch $D_M$ and line $8$ is an optimization step of Eq. \ref{eq:elbo_mini_svgp}. Line $4$ randomly samples a beam $p_b$ collected at time $t_b <= t$ from the MBES log. Line $5$ finds the corresponding particle's pose at $t_b$ and lines $6-7$ transforms $p_b$ to that pose and adds it to the minibatch. Instructions $4-7$ cannot be parallelized since they involve accessing the common MBES log, which is unique due to memory constraints (note that line $8$ is indeed parallel, see Section \ref{subsec:rbpf_structure}). Therefore they need to be very efficient in order to maximize SVGP training iterations across particles. Backtracing through an ancestry tree in line $5$ as in \cite{barkby2011bathymetric, krasnosky2022massively} is therefore unsuitable here. Thus we present a new implementation of the management of particles' trajectories histories that sidesteps the need to explicitly keep track on a tree of the particles' lineages upon resampling. Our C++ approach represents histories $h_t^j$ as vectors of managed pointers to the vectors containing the segments of particles' trajectories in between resamples. Our approach is illustrated in Fig. \ref{fig:particle_mem}.

The left image in Fig. \ref{fig:particle_mem} depicts an example with three particles’ trajectories in 2D and three resampling steps, named $LC_{i}$. In red, the trajectory of the winning particle after each resampling step. On the right, their three corresponding trajectory histories $h_t^j$ as vectors of pointers evolving over time. During the time before the first loop closure, $LC_0$, each particle stores its own trajectory as a vector of poses managed by a pointer. Once $LC_0$ is triggered and particle 1 resampled, the pointers in particle 0 and 2 delete their content and point at the trajectory segment in particle 1. After, a new pointer is added to each history vector and the process continues so that after $LC_2$, each particles' histories either contains or points at the three corresponding segments of trajectories between LCs. Thus, iterating through a particle’s history is trivial provided a common, auxiliary vector containing the length of the trajectory segments (as integers). Given a random beam at time $t_b <= t$, the vector will signal at the corresponding pointer containing the particle’s pose at $t_b$, $r^j_{tb}$.

Additionally, as a result of this implementation computing the final RBPF trajectory can be done directly just by averaging all the particles histories. The need to explicitly keep track of their lineages has been sidestepped. 

\subsection{LC prompting and particle resampling}
Given the set of $J$ particles $R_t$ in Eq. \ref{eq:particles}, the resampling procedure on the RBPF follows the steps in algorithm \ref{algo:resampling}.

\begin{algorithm2e}
\caption{LC prompting} \label{algo:resampling}
\small
    \SetKwFunction{FMain}{lc_prompting}
    \SetKwProg{Fn}{}{:}{}
    \Fn{\FMain{$R_t, P_t, Q$}}{
        $\hat{R_t} = \emptyset$ \\
        \For{\texttt{$\langle r^j_t, SVGP^j_t \rangle $ in $R_t$}}{
            $\hat{P}_t^j = T(r_{t}^j) \: P_t$\\
            $w_t^j = \mathcal{N}(P^{z}_t; SVGP^j_{\mu}(\hat{P}_t^{j, xy}), SVGP^j_{\Sigma}(\hat{P}_t^{j, xy}) + Q)$ \\
            $add$ $\langle r^j_t, w^j_t, SVGP^j_t \rangle$ $to$ $\hat{R}_t$ \\
        }
        $R_t = \emptyset$ \\
        \For{\texttt{$j$ in $J$}}{
            $draw$ $j$ $from$ $\hat{R}_t$ $with$ $probability$ $\propto w_t^j$ \\
            $add$ $\langle r^j_t, SVGP^j_t \rangle$ $to$ $R_t$
        }
     \KwRet{$R_t$}
    }
\end{algorithm2e}

For each particle $j$, a new weight $w_t^j$ is computed in lines $4-6$ comparing the depths of the latest ping $P_t$ against the predictions of that particle's SVGP map. The weight will be proportional to the likelihood of the difference between the real and the expected ping depths being zero. The expected depths can be computed for a given set of 2D query locations $X_* = \hat{P}_t^{j, xy}$ via the posterior of each SVGP from Eq. \ref{eq:bayesian_svgp}, which is given by
\begin{multline}
\label{eq:posterior_svgp}
    p(f(X_{*}) | Y) \sim  \mathcal{N}(K_{*s}K^{-1}_{ss}\mu, \\
    (K_{*s}K^{-1}_{ss})\Sigma(K_{*s}K^{-1}_{ss})^T + K_{**} - K_{*s}K^{-1}_{ss}K^{T}_{*s}))
\end{multline}

If the effective number of particles \cite{liu1996metropolized} is greater than half the number of particles, a systematic resampling \cite{kitagawa1996monte} takes place in lines $8-10$.
The resampling step in line $10$ might require copying particles. In our framework, this process entails two operations: i) copying $h_t^j$, which is extremely fast as it is a vector of pointers and ii) copying the SVGP map. The later involves not only the SVGP hyperparameters $\theta_{SVGP}$, but also the state of the particle's optimizer, so that the training can continue in the new particle.

Finally, to save computational power and avoid GPU congestion, the LC prompting process in algorithm \ref{algo:resampling} occurs with a frequency $f_{LC}$, which is user defined but constrained by the number of particles and the hardware capacity. However, LC prompts are enabled only after every SVGP ELBO has converged, according to the criterion defined in Section \ref{subsec:svgp_training}.

\subsection{The RBPF parallel structure}
\label{subsec:rbpf_structure}
As opposed to a custom-made implementation of the SVGP regression for our RBPF, our solution builds upon GPytorch \cite{gardner2018gpytorch}. The advantages are that GPytorch is open-source, general and community-based and maintained. The trade off rises from the so-called lazy management of the CUDA contexts carried out by Torch. Currently, GPytorch lazily instantiates a new CUDA context (600-800 MB) per SVGP created in a new process. This implies that a fully parallel implementation of the SVGP maps would be quickly limited by memory requirements.
To overcome this, we have adopted a quasi-parallel implementation of the SVGP training. In this solution, a set of $B$ parallel nodes, each with its own CUDA context, hosts each $b$ SVGP maps, which run optimization steps recursively within the node. Thus, the total number of particles per nodes is $b = J/B $. An advantage of this implementation, however, is that the maximum number of SVGP maps accessing the GPU at the same time is now $B$, which can be limited by the user to prevent overloading it.

\begin{figure}[htbp]
    \centering
    \includegraphics[width=0.8\linewidth]{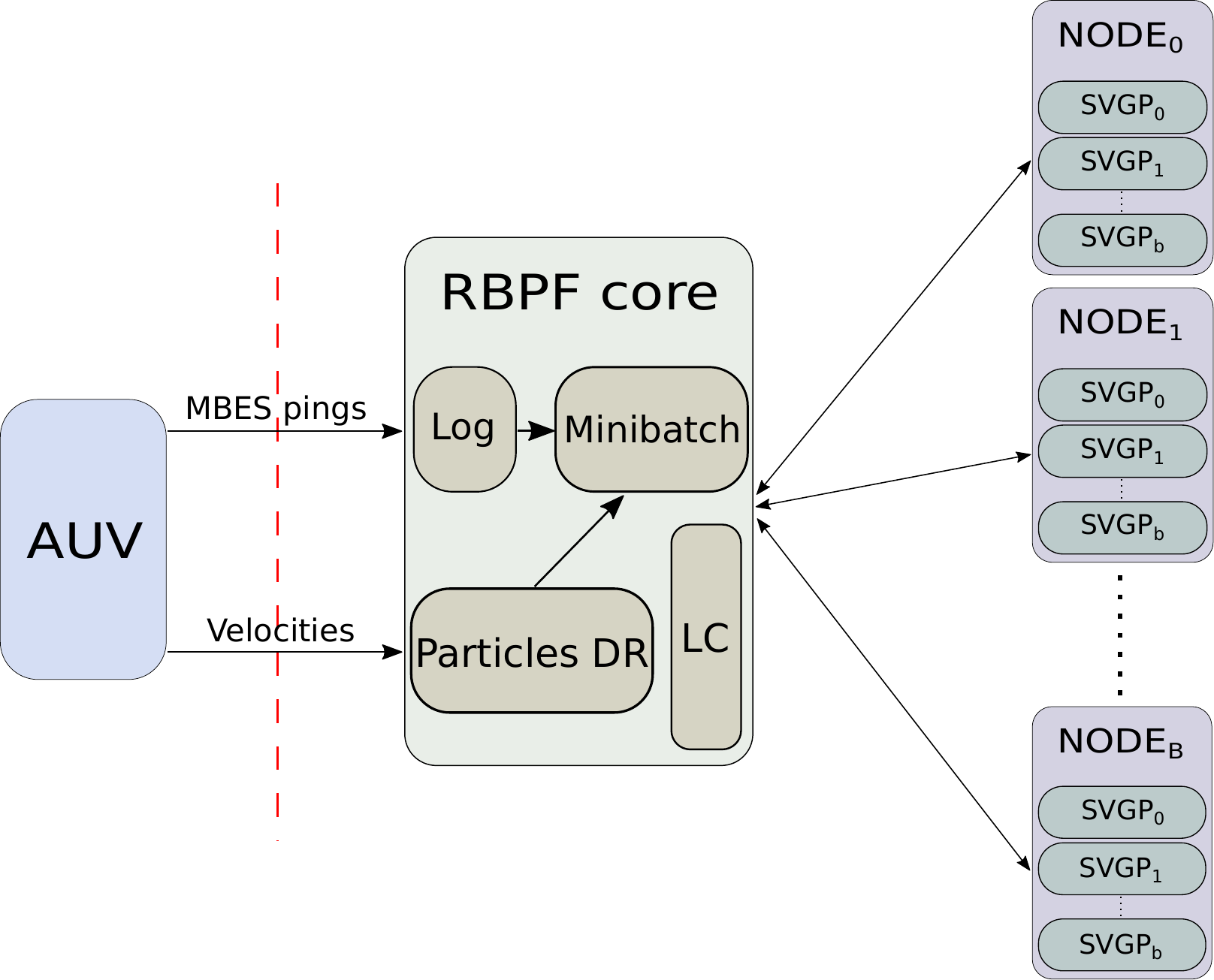}
    \caption{Schematic of the RBPF quasi-parallel structure, implemented in ROS}
    \label{fig:rbpf_schematic}
\end{figure}

Fig. \ref{fig:rbpf_schematic} depicts a simplified graph of the system, implemented in ROS \cite{quigley2009ros}. Each process from the set $O$ is a parallel ROS node, whose SVGP maps implement recursively the key operations of SVGP training (line $8$ on algorithm \ref{algo:minibatch}) and posterior sampling (line $5$ in algorithm \ref{algo:resampling}) when attempting a LC. The RBPF core node is written in C++ and both maintains the MBES log $D_t$ and computes the particles propagation whenever new data from the AUV is received. Furthermore, it stores the particles' histories $H_t = \{ h_t^j\}_{j=0}^{J}$ and keeps them synchronized with $D_t$ for an efficient minibatch construction when required by an SVGP map. Additionally, it handles the LC prompting, the weights computation based on the expected measurements from the SVGP maps and the particles resampling. All these task are implemented in parallel.

\section{EXPERIMENTS}
This section presents the evaluation methods designed and the datasets used to test the performance of the RBPF-SVGP framework proposed.

\subsection{Datasets and vehicles}
Three bathymetric surveys off the Swedish coast have been collected with two AUVs for testing the RBPF. A Kongsberg Hugin 3000 equipped with a MBES Kongsberg 2040 has collected mission $1$ without any external navigation aid. Surveys $2$ and $3$ were executed by the AUV Lolo \cite{deutsch2018design}, with a R2Sonic MBES. While Hugin is a commercially available vehicle with high-end specifications, Lolo is a research platform under development at KTH university and as such it has been a suitable vehicle to test our framework live in mission $2$ in its payload computer, an NVIDIA Jetson Orin. 
\begin{figure}[h]
    \centering
    \includegraphics[width=\linewidth]{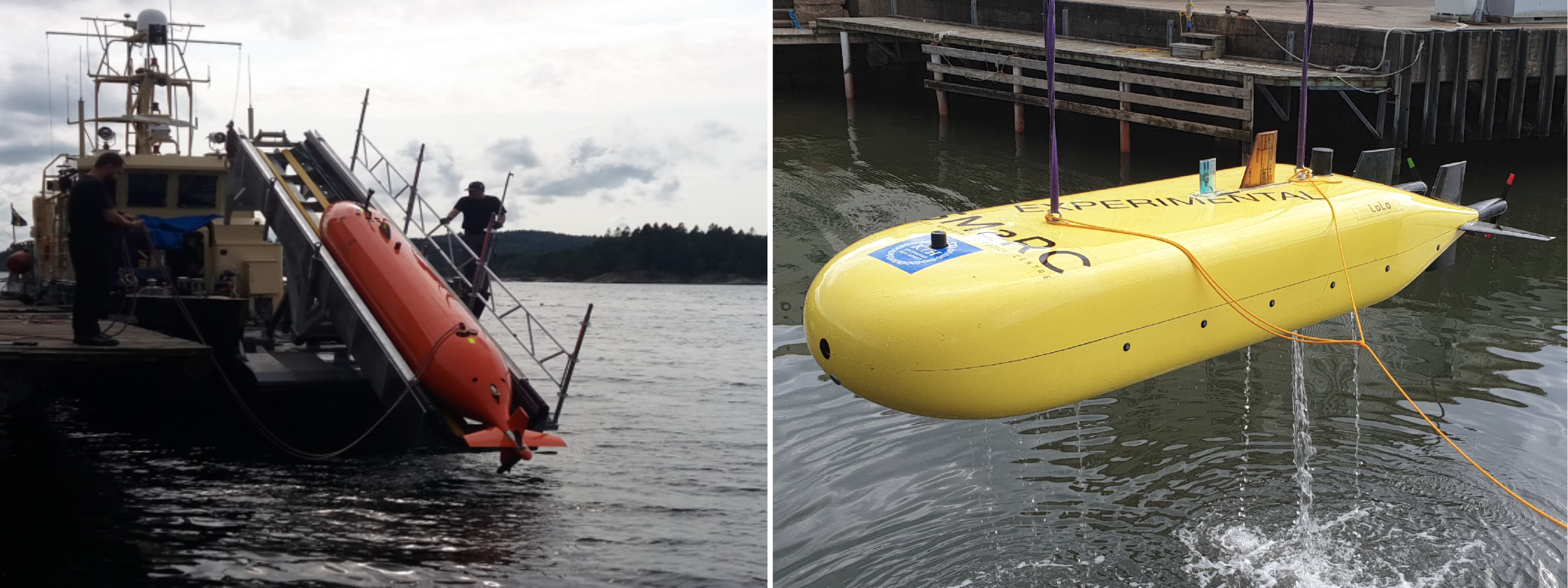}
    \caption{The AUVs Hugin (left) and Lolo (right) being deployed for the surveys in the experiments.}
    \label{fig:auvs}
\end{figure}

The Jetson Orin features a 12-core ARM processor, 32GB of RAM and a 2048-core NVIDIA Ampere architecture GPU with 64 tensor cores. All the experiments have been conducted in the Orin, however only mission 2 was carried out live in Lolo. The experiments on missions $1$ and $3$ were executed offline and therefore the ROS environment from \cite{torroba2022fully} has been used to replay the mission from an external PC acting as the AUV.
Table \ref{tab:missions} summarizes the main characteristics of three missions.

\begin{table}[htbp]
\centering
\vspace*{2mm}
\caption{Parameters for the 3 experiments}
\begin{tabular}{|c|ccc|}
\hline
 Experiment & 1 & 2  & 3  \\ \hline
 AUV    & Hugin & Lolo  & Lolo  \\ 
 Avg velocity (m/s) & 2.0 & 0.75  & 0.6  \\ 
 Duration (h) & 4.45 & 2.25  & 0.83  \\ 
 $N$ (beams) & 8.075.900  & 12.392.200  & 4.658.200  \\ 
 Minibatch size (M)    & 1000  & 1000 &  1000  \\
 Inducing points (S)    & 400  & 400 &  300  \\
 $f_{LC}$           & - & - &  0.1  \\ \hline
\end{tabular}
\label{tab:missions}
\end{table}

\subsection{SVGP training}
\label{subsec:svgp_training}
The Matern $\tau = 1/2$ covariance function has been selected for the SVGP since it has proved to work well on bathymetric data in \cite{torroba2022fully}. The ELBO optimization has been carried out with Adam \cite{kingma2014adam} with a learning rate of $1e^{-1}$ across missions. The size of the minibatch $M$ and the number of inducing points $S$ has been tuned for each mission, depending on the extension of the area to survey. The final parameters can be seen in Table \ref{tab:missions}.

A key requirement for the RBPF resampling to be successful is that the SVGP maps resemble the true bathymetry close enough before attempting loop closures. For this, an exponential moving average is computed over the ELBO of each SVGP \cite{balandat2019botorch} at each optimization step in order to assess convergence at time $t$. However, in our online setup, convergence at time $t$ does not terminate the training, it only signals the RBPF core that the SVGP is ready to be used for LC prompting. This is due to the fact that in online training the SVGP ELBO can temporarily diverge when new data corresponding to previously unseen seabed relief is being added to $D_t$. This motivates continuously training the SVGP throughout the full mission. Fig. \ref{fig:elbo_online_svgp} showcases such phenomena in a particle's SVGP ELBO from experiment 1. 


\begin{figure*}[!t]
\vspace*{0.1in}
\hspace{.1in}(a)\hspace{1.4in}(b)\hspace{1.7in}(c)\hspace{1.5in}(d)\\
\centering
\begin{subfigure}[]{.21\linewidth}
\centering
	\includegraphics[width=\linewidth]{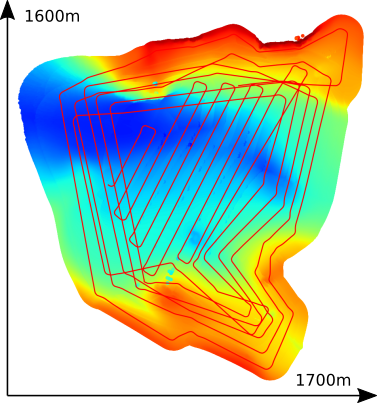}
\label{fig:pcl_map}
\end{subfigure}
\begin{subfigure}[]{.21\linewidth}
\centering
        \includegraphics[width=\linewidth]{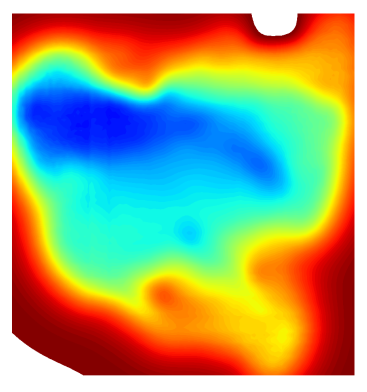}
\label{subfig:disrupted_bathy}
\end{subfigure}
\begin{subfigure}[]{.035\linewidth}
\centering
		\includegraphics[width=\linewidth]{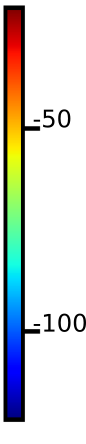}
\end{subfigure}
\begin{subfigure}[]{.21\linewidth}
\centering
		\includegraphics[width=\linewidth]{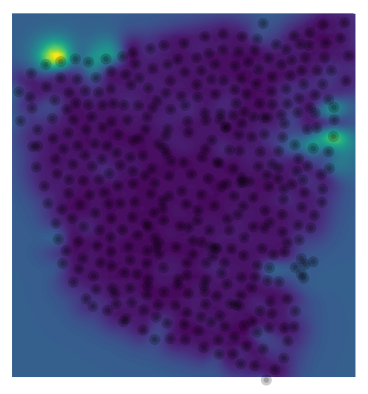}
\label{subfig:gp_ui}
\end{subfigure}
\begin{subfigure}[]{.035\linewidth}
\centering
		\includegraphics[width=\linewidth]{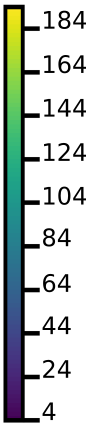}
\end{subfigure}
\begin{subfigure}[]{.21\linewidth}
\centering
		\includegraphics[width=\linewidth]{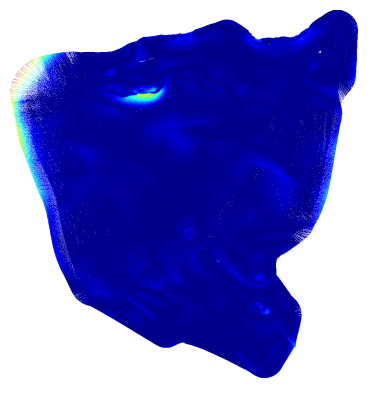}
\label{subfig:rmse_gp_ui}
\end{subfigure}
\begin{subfigure}[]{.028\linewidth}
\centering
		\includegraphics[width=\linewidth]{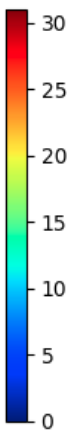}
\end{subfigure}

\begin{subfigure}[]{.21\linewidth}
\centering
	\includegraphics[width=\linewidth]{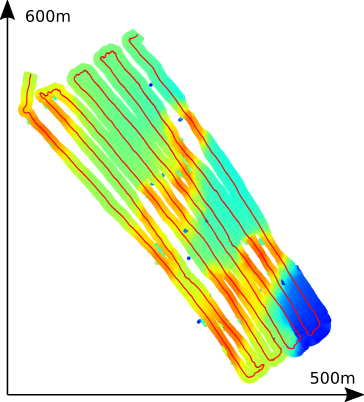}
\label{fig:pcl_map}
\end{subfigure}
\begin{subfigure}[]{.21\linewidth}
\centering
        \includegraphics[width=\linewidth]{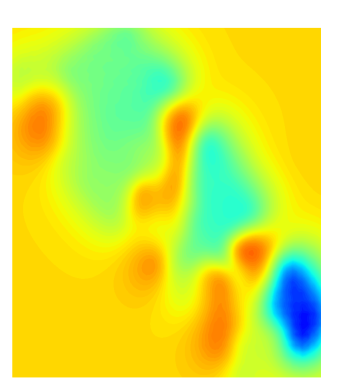}
\label{subfig:disrupted_bathy}
\end{subfigure}
\begin{subfigure}[]{.03\linewidth}
\centering
		\includegraphics[width=\linewidth]{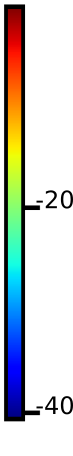}
\end{subfigure}
\begin{subfigure}[]{.215\linewidth}
\centering
		\includegraphics[width=\linewidth]{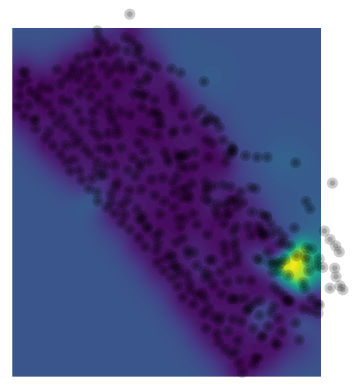}
\label{subfig:gp_ui}
\end{subfigure}
\begin{subfigure}[]{.03\linewidth}
\centering
		\includegraphics[width=\linewidth]{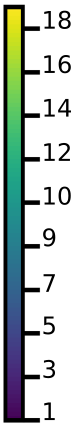}
\end{subfigure}
\begin{subfigure}[]{.21\linewidth}
\centering
		\includegraphics[width=\linewidth]{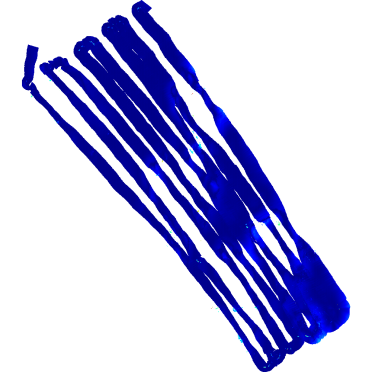}
\label{subfig:rmse_gp_ui}
\end{subfigure}
\begin{subfigure}[]{.028\linewidth}
\centering
		\includegraphics[width=\linewidth]{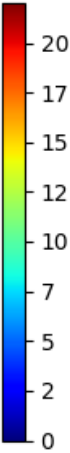}
\end{subfigure}

\caption{Top row: experiments 1 with configuration $J=100$, $B=10$. Columns: a) raw bathymetry and DR trajectory. b) SVGP posterior mean and c) variance and inducing points locations of one particle randomly selected at the end of the experiment. d) Consistency error map from comparing a) and b). Bottom row: experiments 2 live in Lolo. Units are meters.}
\label{fig:gp_maps_resuls}
\end{figure*}

\subsection{Evaluation of the RBPF-SVGP framework}
We now seek to assess the amenability of the proposed framework to real-time SLAM in large seabed areas in an embedded platform. For this, we have designed three sets of experiments carried out in the Jetson Orin. In experiment 1, we have gauged the effects of the parallelization of the SVGP training in the RBPF capacity to train as many SVGP maps as possible in real time during a very large survey with Hugin. We have evaluated the capabilities of the parallel structure proposed in terms of training iterations per particle across the mission and the resulting SVGP maps resemblance to the original bathymetry in terms of the consistency error metric from \cite{roman2006consistency}. Experiment 2 shows the results from testing the same setup in Lolo live during a survey. In experiment 3, we have evaluated the performance of the RBPF in terms of the estimated vehicle trajectory compared against the ground truth computed from GPS measurements \cite{olson2009evaluating} on Lolo data.

\section{RESULTS}
This section presents the results from testing the RBPF-SVGP framework on the terrain modelling and localization tasks on the experiments introduced above.

\subsection{Experiment 1: online SVGP mapping parallelization}
As mentioned in the previous section, for the RBPF to converge towards the true vehicle pose upon LC, the SVGP maps need to resemble the true underlying bathymetry closely enough. This requires that each SVGP achieves enough training iterations in real time, which is the reason behind the highly efficient and parallel implementation of the RBPF presented. In order to measure the resemblance of the SVGP maps with respect to the raw bathymetry, we have tested the mapping-only capability of the SVGP-RBPF with the data from mission 1 and applied the error metric from \cite{roman2006consistency} to assess it. For this type of experiment, the particles have been configured to follow the vehicles' DR, with zero motion noise added and the LC prompting turned off.

\begin{table}[htbp]
\centering
\caption{Results from the parallel mapping-only experiments with the Hugin survey}
\begin{tabular}{l|llll|}
\cline{2-5}
& \multicolumn{4}{c|}{Experiment 1} \\ \hline

\multicolumn{1}{|l|}{Particles (J)}    & \multicolumn{1}{c|}{1} & \multicolumn{1}{c|}{10} & \multicolumn{1}{c|}{10} & \multicolumn{1}{c|}{100}            \\

\multicolumn{1}{|l|}{Nodes (B)}        & \multicolumn{1}{c|}{1} & \multicolumn{1}{c|}{1} & \multicolumn{1}{c|}{10} & \multicolumn{1}{c|}{10}              \\ \hline

\multicolumn{1}{|l|}{Avg it/particle} & 202.327 & 21.425 & 39.373 & 3.714   \\


\multicolumn{1}{|l|}{Map RMSE} & 1,636 & 1,709 & 1,712 & 1,995 \\ \hline

\end{tabular}
\label{tab:map_exp}
\end{table}

The first row of images in Fig. \ref{fig:gp_maps_resuls} shows the results from this experiment with the configuration $J=100$, $B=10$ for one particle randomly picked at the end of the mission. Column a) depicts the original bathymetry map collected by Hugin with the DR trajectories in red. Columns b) and c) represent the final particle's SVGP posterior mean and variance. The final locations of the inducing points are plotted over the variance map. Column d) illustrates the consistency error map from comparing a) and b) using the error metric from \cite{roman2006consistency}. Notice that the peak errors correspond to either bathymetry areas with sharp changes, known to be difficult to model with GPR, or to noise in the MBES data. 

To gauge the effects of the parallelization of the SVGP training, the Hugin survey has been replayed with four particles configurations, summarized in Table \ref{tab:map_exp}. Namely: single particle ($J=1$, $B=1$), no parallelization ($J=10$, $B=1$), full parallelization ($J=10$, $B=10$) and quasi-parallelization ($J=100$, $B=10$). Notice how the fully parallel configuration has achieved roughly double the total amount of iterations in the mission than the fully sequential. Given that a $100 \%$ utilization of the GPU was achieved, we believe $\sim 400.000$ to be the maximum number of SVGP training iterations achievable in this hardware with the proposed framework, regardless of the degree of parallelization. Approximately the same amount has been achieved by the quasi-parallel implementation for a $\times 10$ larger number of particles, indicating maximum utilization of the GPU as well.

Regarding the resulting SVGP map, the fully-parallel configuration has achieved the same consistency RMS error than the sequential one for double the number of iterations per particle, which signals overfitting of the maps. We therefore adopt the $RMSE \sim 1,7m$ as the baseline and we see how a tenfold increment of the particles only causes an increase of $17,35 \%$ on the error, which hints at a good performance of our quasi-parallel implementation.

\begin{figure}[h]
    \centering
    \vspace*{2mm}
    \includegraphics[width=\linewidth]{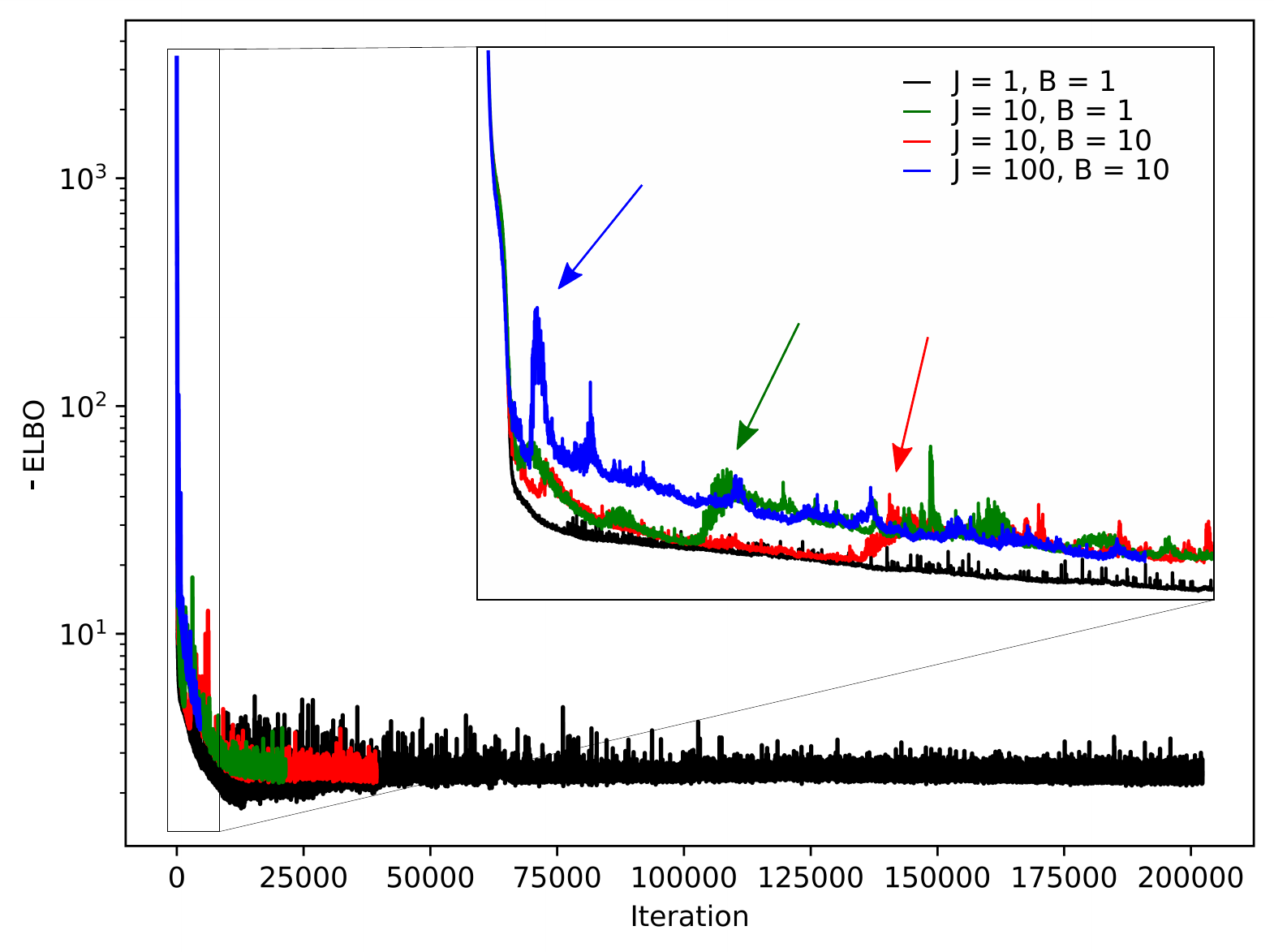}
    \caption{Resulting -ELBO of one particle for the four configurations in experiment 1. Zoom-in: notice the effect of the full parallelization (red) compared to the fully sequential training (green) on the shape of the bump marked with arrows. }
    \label{fig:elbo_online_svgp}
\end{figure}
Finally, concerning the influence of the filter configuration in the temporal aspect of the online training, Fig. \ref{fig:elbo_online_svgp} illustrates the evolution of the ELBO of a particle randomly sampled from each configuration in Experiment 1. The initial bump caused by the AUV entering the flat area of the survey in Fig. \ref{fig:gp_maps_resuls} a) row 1 has been marked with arrows. As expected, it is smoother the more SVGP iterations are attainable per MBES ping collected. To enable configurations with large number of particles, the LC prompting is only enabled once a certain degree of convergence has been achieved by all the SVGP maps, as explained in \ref{subsec:svgp_training}. Thus, the initial bumps can be safely sidestepped.

\subsection{Experiment 2: live survey in Lolo}
The same setup as in experiment 1 has been tested live in Lolo while surveying the area in the second row of Fig. \ref{fig:gp_maps_resuls}, column a). Due to technical issues in Lolo at the time of the data collection, the mission was carried out on the surface. For this experiment, the particles configuration has been set to $J=80, B=8$ to achieve a better training performance without overloading the CPUs. The CPUs on the Jetson payload needed to be shared among the mission behaviour tree \cite{sprague2018improving} and other secondary processes. Therefore, although more particles could have been set for the experiment, it would have been at the expense of a lower performance of the GPU. The final average number of iterations per particle was $2965$ and the consistency RMS error $1,003$.
The output of the live mapping for a particle randomly chosen can be seen in the second row of images in Fig. \ref{fig:gp_maps_resuls}, columns b)-d).

\subsection{Experiment 3: RBPF localization estimate}
\begin{figure}[h]
    \centering
    \includegraphics[width=\linewidth]{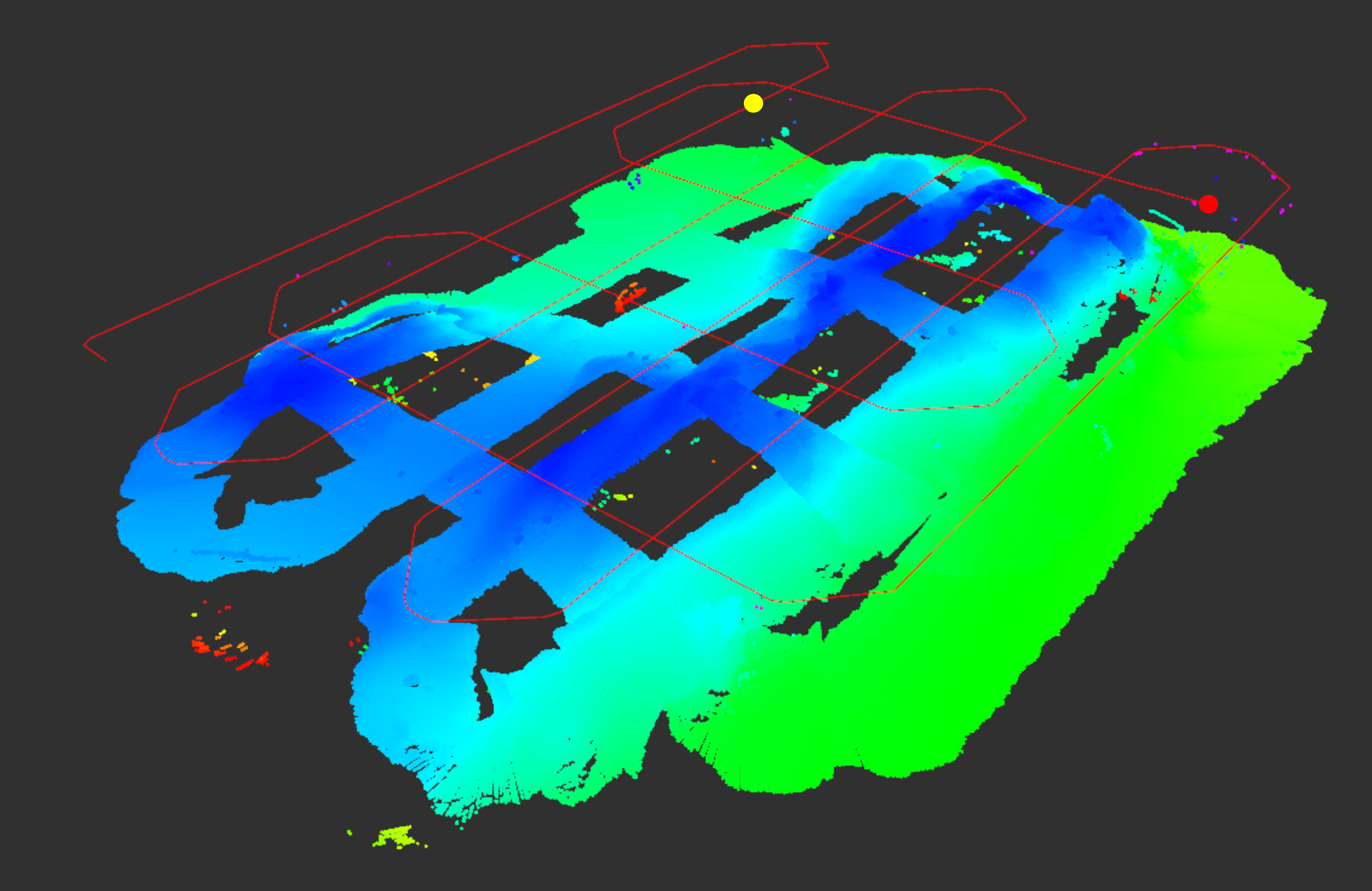}
    \caption{RViz visualization of the raw bathymetry and the planned trajectory (red) on mission 3. The yellow marker signals the starting point (the first swath was discarded) and the red one, the end of the mission.}
    \label{fig:rbpf_traj}
\end{figure}

To assess the overall performance of the RBPF framework in a SLAM problem, the full RBPF-SVGP algorithm is tested in survey 3 with $J = 100$ and $B = 10$. This is a challenging survey collected with Lolo that presents large coverage gaps in the bathymetry and strong noise in the DR and the MBES, as can be seen in Fig. \ref{fig:rbpf_traj} The area surveyed is approximately $250 \times 150$ m. 
This figure depicts the planned trajectory in red, with a yellow marker signalling the starting point of the mission (the first swath was neglected) and a red one the end. As in the previous experiment, due to technical problems with Lolo the mission was carried out on the surface. This, however, has allowed collecting GPS measurements that have been utilized as the ground truth trajectory. Nevertheless, the AUV's DR used in the experiment has been estimated only from the velocities provided by the navigation system. 
\begin{figure}[h]
    \centering
    \includegraphics[width=\linewidth]{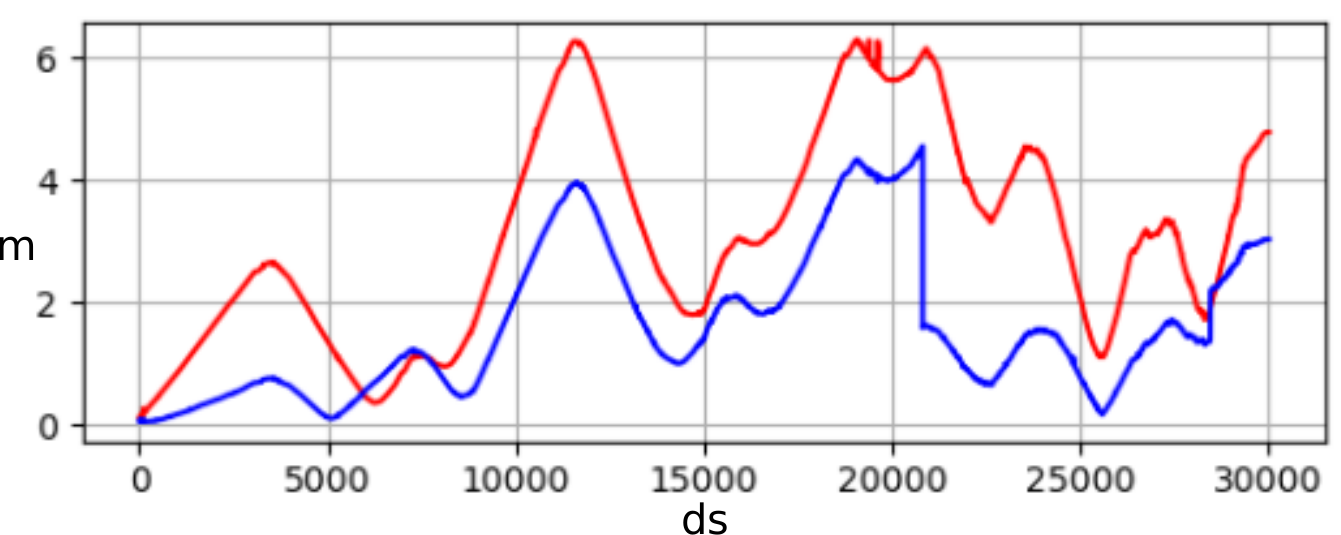}
    \caption{DR (red) and RBPF (blue) trajectory errors against GT during experiment 3. Notice the two loop closures at $t_{step} \sim 21000$ and $t_{step} \sim 28000$.}
    \label{fig:rbpf_error}
\end{figure}

Fig. \ref{fig:rbpf_error} depicts the evolution over time of the DR (red) position error, in meters, against the GT position (from the GPS). In blue, the RBPF trajectory estimate during the mission, computed online from the average of the particles. The RBPF has registered two loop closures: a correct one at $t_{step} \sim 21000$ and an erroneous one at $t_{step} \sim 28000$. The high level of spurious measurements in the MBES data has led us to manually set the uncertainty of the MBES noise model ($Q$) to a high level in order to make the filter more robust against outliers, however resulting in only two loop closures detected.

\section{CONCLUSIONS AND FUTURE WORK}
We have presented a Rao-Blackwellized particle filter with Stochastic Variational Gaussian Process maps for online, large-scale SLAM in an embedded platform. The key contributions proposed that have enabled our approach are: i) the modifications of the SVGP training process, which have permitted to adapt SVGP to online regression on surveys with a mobile platform, ii) a new configuration of the RBPF particles' lineage tracking, which has allowed parallel, real-time SVGP training in a platform with limited resources. 

The proposed framework has been implemented and tested in a Jetson Orin and its performance has been assessed on the tasks of parallel, real-time mapping and localization in three challenging surveys collected with two AUVs, one of them live. When compared to the current state of the art solution for underwater RBPF SLAM from \cite{krasnosky2022massively}, our method is capable of handling larger surveys, both in terms of area covered and mission time, with a single SVGP per particle map and with larger sets of particles.
Regarding its implementation, the SVGP regression module of our framework builds upon the open-source GPytorch library and it does not require any modification of the original SVGP formulation, facilitating its use. Additionally, the full framework has been released to encourage further research on the topic \footnote{\url{https://github.com/ignaciotb/UWExploration/}}.

The current major limitation of the implementation concerns the lazy initialization of the CUDA contexts per SVGP model carried out by Pytorch. This limits the number of fully parallel SVGP maps by the amount of available RAM of the host device. Although a quasi-parallel configuration has been proposed, overcoming this constraint would entail a full SVGP implementation in native CUDA. Nevertheless, when enough RAM is available, a fully-parallel setup can be achieved setting $B = J$. Additionally, we propose as further research the use of the particles' ELBO value within the particles weighting process. It has been shown that the ELBO relates to the online state of the SVGP map and therefore factoring it in the particle weighting could potentially result in more accurate resampling.

Finally, although this work has targeted bathymetric SLAM, the proposed approached is not limited to it and it can be applied to different problems and domains.

\section*{ACKNOWLEDGMENT}
The authors thank the Alice Wallenberg foundation for funding MUST, Mobile Underwater System Tools, project that provided the Hugin AUV.

\bibliography{main}
\bibliographystyle{unsrt}

\end{document}